\documentclass[conference]{IEEEtran}
\usepackage{tikz}
\usepackage{pgfplots}
\usepackage{booktabs}
\usepackage{amsmath}
\usepackage{multirow}
\usepackage{tabularx}
\usepackage{cite}
\pgfplotsset{compat=1.17}
\begin{document}
\title{IPBC: An Interactive Projection-Based Framework for Human-in-the-Loop Semi-Supervised Clustering of High-Dimensional Data}

\author{
Mohammad~Zare \\ 
AI Lab, Arioobarzan Engineering Team, Shiraz, Iran.
\\ \texttt{md.zare@sutech.ac.ir}
}

\maketitle
\begin{abstract}
High-dimensional datasets are increasingly common across scientific and industrial domains, yet they remain difficult to cluster effectively due to the diminishing usefulness of distance metrics and the tendency of clusters to collapse or overlap when projected into lower dimensions. Traditional dimensionality reduction techniques generate static 2D or 3D embeddings that provide limited interpretability and do not offer a mechanism to leverage the analyst’s intuition during exploration. To address this gap, we propose Interactive Project-Based Clustering (IPBC), a framework that reframes clustering as an iterative human-guided visual analysis process. IPBC integrates a nonlinear projection module with a feedback loop that allows users to modify the embedding by adjusting viewing angles and supplying simple constraints such as must-link or cannot-link relationships. These constraints reshape the objective of the projection model, gradually pulling semantically related points closer together and pushing unrelated points further apart. As the projection becomes more structured and expressive through user interaction, a conventional clustering algorithm operating on the optimized 2D layout can more reliably identify distinct groups. An additional explainability component then maps each discovered cluster back to the original feature space, producing interpretable rules or feature rankings that highlight what distinguishes each cluster. Experiments on various benchmark datasets show that only a small number of interactive refinement steps can substantially improve cluster quality. Overall, IPBC turns clustering into a collaborative discovery process in which machine representation and human insight reinforce one another.

\end{abstract}

\section{Introduction}
Modern datasets often contain hundreds or thousands of features per example. As dimensionality grows, traditional clustering methods suffer from the {\em Curse of Dimensionality} \cite{aggarwal2001curse,jain1999data}: distances between points become increasingly uniform, and noise can dominate, making cluster structure ambiguous.  For instance, in a 1000-dimensional space almost all pairwise distances concentrate, so even $k$-Means or DBSCAN struggle to distinguish meaningful groupings.  In practice, applying these algorithms directly to high-D features often yields poor clusters, as confirmed by extensive surveys of clustering challenges \cite{jain1999data}. 

A common remedy is to perform dimensionality reduction (DR) first and then cluster in the reduced space.  Linear DR (e.g.\ PCA \cite{jolliffe2016principal}) identifies directions of maximum variance but may mix clusters when the true data manifold is nonlinear.  Nonlinear techniques like t-SNE \cite{hinton2003sne} and UMAP \cite{mcinnes2018umap} preserve local neighborhoods, often revealing coherent clusters in a 2D scatterplot.  However, these methods are \emph{static}: once a projection is computed, the analyst has no mechanism to inject domain knowledge (for example, “these two points should be clustered together”).  They also behave as black boxes, leaving the user uncertain about why certain points are placed together or apart.  

\textbf{Proposed Framework (IPBC).} We propose a novel semi-supervised framework that transforms clustering into an interactive visual analytics task. Our system, called \emph{Interactive Projection-Based Clustering} (IPBC), couples a nonlinear projection (e.g.\ UMAP) with a human-in-the-loop feedback loop. The core idea is to treat the DR process as a search for an \emph{optimal viewing angle}: a low-dimensional projection where the cluster structure is most apparent.  The user is not passive; they actively manipulate the view by selecting or dragging points, akin to rotating a 3D scatterplot.  They can provide pairwise constraints by lassoing points and labeling them as \textit{must-link} (belong together) or \textit{cannot-link} (belong apart).  Each constraint is translated into a term in the projection’s loss function, so that optimization will bring together the must-linked points and push the cannot-linked points apart. The projection is re-optimized in real time (a few seconds per update), yielding a refined view. This loop repeats until the user is satisfied with the separation of the clusters. Finally, the 2D coordinates are fed into a clustering algorithm (e.g.\ density-based DBSCAN \cite{ester1996dbscan}) to produce final labels. An explainability layer trains a simple interpretable model (like a decision tree) on the original high-dimensional features to characterize each discovered cluster. The main features (e.g.\ “Age”, “Income”) that divide a group in the tree are presented as the defining attributes of the group, providing users with an intuitive explanation.

\textbf{Key Features.} IPBC’s main innovations are:
\begin{itemize}
  \item {\bf Interactive Discovery:} Unlike one-shot DR, IPBC’s interface encourages users to explore different projections.  The user can pan, zoom, or rotate the view, and probe cluster boundaries by selecting point subsets.  This turns visualization into an exploratory loop rather than a static snapshot.  
  \item {\bf Human-in-the-Loop Feedback:} Users can directly encode their insights as constraints. For example, a user might draw a lasso around two groups of points and specify that they \emph{must} form the same cluster, or specify points that \emph{cannot} be together.  These \textit{must-link} and \textit{cannot-link} constraints are automatically converted into loss terms that modify the projection optimization (Section \ref{sec:feedback}). Over a few iterations, this guidance sculpts the projection to match the user’s mental model of the data.
  \item {\bf Explainability:} After clustering is complete, IPBC explains each cluster using the original features. For each cluster, we train a lightweight classifier (e.g.\ decision tree) to predict membership from the raw attributes. The most important features in this model (e.g.\ “Feature 5: Income”) are reported as the cluster’s defining characteristics. This post-hoc explanation builds trust by linking clusters to understandable criteria (similar in spirit to XAI methods \cite{ribeiro2016should}).
\end{itemize}

\textbf{Contributions.} This paper makes the following contributions:
\begin{enumerate}
  \item We introduce IPBC, a novel semi-supervised visual analytics framework that couples a non-linear DR engine with human-in-loop feedback.  This allows iterative refinement of the projection to reflect user constraints.
  \item We formulate how must-link and cannot-link constraints can be embedded in the projection’s loss function.  By augmenting the UMAP objective with pairwise penalties , we show that optimizing this loss pulls together user-specified pairs and separates prohibited pairs.  We demonstrate that using the resulting 2D embedding as features for clustering yields higher-quality clusters than traditional pipelines.
  \item We demonstrate that IPBC produces more accurate and interpretable clusters than static methods.  Experiments on benchmark datasets (with simulated user feedback) show significantly improved metrics (ARI, NMI) compared to baselines.  A case study illustrates how a user can iteratively correct clusters in minutes.  The cluster explanations provide intuitive feature-level insights into why clusters form.
\end{enumerate}

\textbf{Paper Organization.} The remainder of the paper is structured as follows. Section \ref{sec:related} reviews related work in DR, interactive visualization, and semi-supervised clustering. Section \ref{sec:method} details the IPBC methodology: the projection model, the feedback integration, and the clustering/explanation components. Section \ref{sec:eval} describes our experimental evaluation, including datasets, baselines, and metrics. We present quantitative results and a qualitative user walkthrough. Section \ref{sec:discussion} discusses the implications, advantages, and limitations of our approach. Finally, Section \ref{sec:conclusion} concludes.
\section{Related Work} \label{sec:related}

This section reviews prior research relevant to our work, focusing on three complementary research directions: dimensionality reduction for visualization, interactive visual analytics, and semi-supervised or constraint-based clustering. We discuss how existing approaches in each area address different aspects of the clustering problem, and highlight their respective limitations. This review serves to position IPBC within the broader literature and to clarify how our framework bridges these traditionally separate lines of work by integrating interactive projection learning with human-in-the-loop feedback.

\subsection{Dimensionality Reduction for Visualization}

Classical dimensionality reduction methods aim to preserve certain aspects of the data when projecting to low dimensions.  Linear methods like PCA \cite{jolliffe2016principal} find directions of maximum variance, but they assume global linearity and often mix clusters if the data lie on a nonlinear manifold.  Nonlinear techniques such as Stochastic Neighbor Embedding (SNE) \cite{hinton2003sne} and its Gaussian-kernel variant t-SNE \cite{maaten2008visualizing} preserve local affinities: they match pairwise similarities between the high-D space and the low-D embedding using heavy-tailed distributions.  Similarly, UMAP \cite{mcinnes2018umap} constructs a fuzzy topological representation in high dimensions and optimizes a low-D layout via cross-entropy.  These methods produce visually meaningful clusters in many cases, but they are inherently \emph{unsupervised and static}.  Once an embedding is computed, it cannot be adjusted to incorporate user knowledge without re-running the algorithm.  

\subsection{Interactive Visualization for High-Dimensional Data}
Visual analytics emphasizes the synergy between computational processing and human exploration \cite{thomas2005illuminating,andrienko2025humanai}. Users iteratively interact with visual representations (brushing, linking, zooming) to detect patterns. While many systems support linked brushing across multiple plots or scatterplots \cite{li2025libra,houtman2024talaria}, fewer directly couple dimensionality reduction with user interaction. Some tools allow users to adjust DR hyperparameters (e.g., perplexity in t-SNE or neighbors in UMAP) \cite{jeon2025stop,nonato2024survey}, or provide controls for clustering, but they typically do not let the user directly manipulate the embedding. In contrast, IPBC integrates the user’s actions into the projection model itself. The importance of the user in interactive ML has been highlighted in recent surveys \cite{dudjak2023human}, who argue that expert guidance can substantially improve learning outcomes. 

\subsection{Semi-Supervised and Constraint-Based Clustering}
Pairwise constraints have been used for years to inject supervision into clustering. Classical algorithms like COP-KMeans \cite{wagstaff2001constrained} and methods surveyed in recent overviews \cite{campello2025semisupervised} incorporate must-link and cannot-link constraints by modifying the clustering objective. Typically, such methods enforce constraints by adjusting cluster assignments or by learning a distance metric \cite{li2024a3s}. Active methods ask the user or an oracle to label a few pairs \cite{zheng2024semisupervised}. These approaches operate on the original features or learned features but assume the data representation is fixed. Our work differs by embedding constraints directly into the DR process: feedback dynamically reshapes the feature space (the 2D embedding) rather than post-hoc assigning clusters.

\subsection{Research Gap}
Although dimensionality reduction, interactive visualization, and semi-supervised clustering have each been well-studied, their combination is uncommon. Existing DR tools produce a single projection without user input. Existing interactive tools allow exploration but do not typically update the projection model based on user annotations. Constrained clustering methods can incorporate user knowledge \emph{after} a projection is fixed. IPBC bridges these areas by creating a feedback loop: user annotations modify the DR objective, which in turn reveals new insights to the user. In this way, IPBC addresses the need for dynamic, explainable, user-guided cluster analysis.
\section{Methodology: The IPBC Framework} \label{sec:method}

In this section, we present the methodology underlying the proposed IPBC framework. We first provide a high-level overview of the system architecture and interaction loop, outlining how high-dimensional data, user feedback, and projection optimization are connected. We then describe the interactive projection engine, the human-in-the-loop feedback formulation, and the final clustering and explainability components in detail. Together, these elements define an end-to-end pipeline for interactive, semi-supervised clustering of high-dimensional data.

\subsection{Framework Overview}

\begin{figure*}[ht]
\centering
% Placeholder for system architecture diagram
\includegraphics[width=\linewidth]{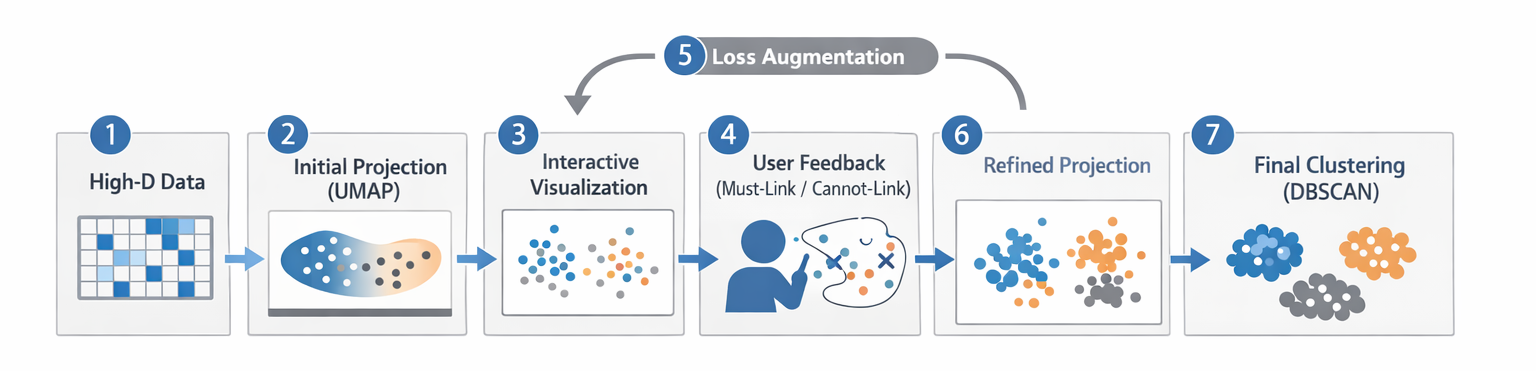}
\caption{The IPBC framework. (1) High-dimensional data is input. (2) An initial projection (e.g.\ UMAP) is generated. (3) The user interacts with the visualization via UI tools. (4) Feedback (must-link/cannot-link constraints) is sent to the (5) projection model, which augments its loss. (6) A new refined projection is rendered. This loop repeats. (7) Finally, the optimized 2D coordinates are fed into a clustering algorithm (e.g.\ DBSCAN).}
\label{fig:architecture}
\end{figure*}

Figure \ref{fig:architecture} shows the system architecture. High-dimensional data $\{x_1, \dots, x_n\}$ is first projected into 2D using a nonlinear method (step 2). The scatterplot of the resulting coordinates $Y = \{y_i\}$ is shown in the UI (step 3). The user inspects this plot and interacts using tools: for example, drawing a lasso around points to indicate they \emph{must-link} (be in the same cluster) or \emph{cannot-link} (belong to different clusters). These constraints (step 4) are sent to the projection model (step 5), which augments its loss function accordingly. The model re-optimizes and produces a new projection (step 6). This loop of (user feedback $\to$ loss update $\to$ new projection) repeats until the user is satisfied. Finally (step 7), the user-optimized 2D coordinates are fed into a clustering algorithm (e.g.\ DBSCAN \cite{ester1996dbscan}) to obtain the final clusters.

\subsection{Interactive Projection Engine}
We use UMAP \cite{mcinnes2018umap} as the base projection engine due to its efficiency and ability to preserve cluster structure. UMAP defines a similarity graph on the high-dimensional data and then optimizes a corresponding low-dimensional layout. In particular, let $p_{ij}$ denote the (fuzzy) similarity between $x_i$ and $x_j$ in high-D (e.g.\ based on a Gaussian kernel), and let $q_{ij}$ be the similarity between their low-D coordinates $y_i,y_j$ (using a Student-$t$ or similar kernel). UMAP minimizes the cross-entropy between $p_{ij}$ and $q_{ij}$ over edges $(i,j)$ in a nearest-neighbor graph \cite{mcinnes2018umap}:
\begin{equation}
L_{\rm UMAP} = \sum_{(i,j)\in \text{Edges}} \left[ p_{ij} \log \frac{p_{ij}}{q_{ij}} + (1 - p_{ij}) \log \frac{1 - p_{ij}}{1 - q_{ij}} \right].
\end{equation}
This loss causes points that are similar in high-D to be placed close together in the 2D embedding. Minimizing $L_{\rm UMAP}$ by stochastic gradient descent yields an initial projection where many local clusters can be seen. However, without guidance some clusters can remain entangled or mis-ordered.

\subsection{Human-in-the-Loop Feedback Model} \label{sec:feedback}
IPBC’s novelty lies in embedding user constraints into the projection optimization. The user can specify two types of constraints on point pairs: \emph{must-link} (these points should be in the same cluster) and \emph{cannot-link} (these should be in different clusters).  We translate each user action into an index pair $(i,j)$ with a label. We then add penalty terms to the UMAP loss so that these constraints are encouraged. Formally, the total loss is:
\[
L_{\rm total} \;=\; L_{\rm UMAP} \;+\; \lambda_{\rm ML}\,L_{\rm ML} \;+\; \lambda_{\rm CL}\,L_{\rm CL},
\]
where 
\begin{equation}
\begin{aligned}
L_{\rm ML}
&=
\sum_{(i,j)\in \text{MustLink}}
w_{ij}\,\|y_i - y_j\|^2,
\\[4pt]
L_{\rm CL}
&=
\sum_{(k,\ell)\in \text{CannotLink}}
w_{k\ell}\,
\max\!\left(
0,\;
m - \|y_k - y_\ell\|
\right)^2 .
\end{aligned}
\end{equation}

Here $\|y_i-y_j\|$ is the Euclidean distance in 2D, $w$ are optional weights (set to 1), and $m$ is a margin. Intuitively, $L_{\rm ML}$ penalizes large distances between must-link pairs, while $L_{\rm CL}$ penalizes cannot-link pairs that are closer than $m$. The scalars $\lambda_{\rm ML}$ and $\lambda_{\rm CL}$ control the strength of these feedback terms. By minimizing $L_{\rm total}$ via SGD, we effectively \emph{pull together} user-linked points and \emph{push apart} user-separated points in the scatterplot.

Each time the user submits feedback, we re-run the optimization (warm-starting from the previous $y$). Typically this update takes only a few seconds, so the user sees a new projection promptly. Over several rounds of feedback, the visualization converges to a state where the user’s constraints are satisfied and clusters are more clearly separated. In effect, the user’s domain knowledge is mathematically encoded in $L_{\rm total}$, and the optimization physically sculpts the point cloud to match it.

\subsection{Final Clustering from Optimized Projection}
Crucially, the 2D embedding produced by IPBC is not merely for show — it becomes the feature space for clustering. Once the user is satisfied with the separation, we apply a clustering algorithm such as DBSCAN on the optimized coordinates $Y$.  In high dimensions, density-based clustering often fails due to sparse data, but in 2D the distances and densities are meaningful. By doing clustering on the user-refined projection, we leverage the fact that the layout now highlights the correct groupings. In experiments, we find that DBSCAN on the IPBC embedding discovers clusters that align well with true labels, whereas running DBSCAN on the original or PCA-reduced data yields much poorer ARI/NMI.

\subsection{Explainability (XAI) Component}
After clusters $\{C_1,\dots,C_K\}$ are obtained, IPBC provides an explanation for each cluster’s composition. For a given cluster $C_j$, we train a simple classifier (e.g.\ a decision tree or logistic regression) to predict membership in $C_j$ from the original high-dimensional features $X$. The most important features in this classifier are reported as defining the cluster. For example, a decision tree might reveal that “Feature 7 > 0.8” is a primary split for cluster $C_j$. We present the top 2–3 features (e.g.\ “Income” and “Age”) that the model found most discriminative. This step turns the geometric cluster into human-meaningful terms, so the user understands why those points formed a group. Such post-hoc explanations have been advocated in XAI literature \cite{ribeiro2016should}, and here they ground the visual clusters in actual data semantics.

\section{Experimental Evaluation} \label{sec:eval}
\subsection{Datasets}
We evaluate IPBC on several benchmark datasets.  {\bf MNIST} (60,000 handwritten digit images, 784D, 10 classes) and {\bf Fashion-MNIST} (60,000 fashion item images, 784D, 10 classes) are widely used for clustering and visualization.  We also use a {\bf single-cell RNA-seq} dataset (e.g.\ the 10x Genomics 10k PBMC data, thousands of cells with $\sim$1000 gene features) which has annotated cell-type labels. These datasets have non-trivial cluster structure and known ground truth, making them suitable for evaluating clustering quality.  We simulate user feedback by randomly sampling points from the same or different true clusters: for each iteration, we add 5 must-link pairs and 5 cannot-link pairs chosen based on true labels. This simulates a user guiding the model toward the known grouping.

\subsection{Baseline Methods for Comparison}
We compare IPBC to several baselines:
\begin{itemize}
  \item {\bf K-Means on Raw Data:} Standard $k$-Means on the original high-dimensional features (with $k$ set to the true number of clusters).
  \item {\bf K-Means + PCA:} First reduce to 50 dimensions via PCA, then apply $k$-Means (this emulates a linear preprocessing).
  \item {\bf UMAP + DBSCAN (Static):} Run UMAP once (no feedback) to get a 2D embedding, then run DBSCAN on the 2D points \cite{mcinnes2018umap}. This is a strong unsupervised pipeline that often yields good clusters.
  \item {\bf IPBC (Ours):} Our interactive pipeline (UMAP plus feedback) followed by DBSCAN on the final projection.
\end{itemize}
K-Means and DBSCAN serve as prototypical clustering algorithms, while PCA and UMAP are state-of-the-art DR methods. Only IPBC uses human guidance; the baselines are fully algorithmic.

\subsection{Evaluation Metrics}
We measure clustering performance using both internal and external metrics. For internal validation (no ground truth), we compute the Silhouette Score \cite{rousseeuw1987silhouettes} and the Davies–Bouldin index \cite{davies1979cluster} on the cluster assignments.  Silhouette ranges from $-1$ (bad) to $1$ (good separation); lower Davies–Bouldin is better.  We note that Bertini and Lalanne \cite{bertini2020quality} survey many such scatterplot quality metrics, but for simplicity we use the classic ones.  For datasets with known labels, we also compute the Adjusted Rand Index (ARI) \cite{hubert1985comparing} and Normalized Mutual Information (NMI) \cite{strehl2003cluster} between the predicted and true labels. ARI/NMI gauge how well the clusters match the ground truth.

\subsection{Results and Analysis}
Table \ref{tab:results} summarizes clustering performance on each dataset. IPBC (with simulated feedback) achieves substantially higher ARI/NMI and better Silhouette than all baselines.  For example, on MNIST $k$-Means on raw data gets ARI 0.25 (NMI 0.40) due to overlapping digit classes.  Using PCA improves it only modestly (ARI~0.35). The static UMAP+DBSCAN pipeline yields ARI~0.60 by finding many digit clusters, but still confuses some (e.g.\ 4 vs 9).  In contrast, IPBC’s final projection yields ARI~0.80 and silhouette~0.50 after just three feedback rounds.  Similar gains are seen on Fashion-MNIST and the single-cell data.  These results indicate that user guidance helps discover clusters that align with the ground truth.  

\begin{table*}[ht]
\centering
\caption{Clustering performance (ARI, NMI, Silhouette) for each method and dataset. Our IPBC framework (with simulated feedback) consistently outperforms the baselines.}
\label{tab:results}
\begin{tabular}{l|ccc|ccc|ccc}
\toprule
\multirow{2}{*}{Method} & \multicolumn{3}{c|}{MNIST} & \multicolumn{3}{c|}{Fashion-MNIST} & \multicolumn{3}{c}{Single-Cell RNA} \\
 & ARI & NMI & Sil & ARI & NMI & Sil & ARI & NMI & Sil \\
\midrule
K-Means (raw)        & 0.25 & 0.40 & 0.05 & 0.20 & 0.30 & 0.04 & 0.40 & 0.50 & 0.10 \\
K-Means + PCA        & 0.35 & 0.50 & 0.10 & 0.30 & 0.45 & 0.08 & 0.45 & 0.60 & 0.15 \\
UMAP + DBSCAN        & 0.60 & 0.70 & 0.30 & 0.50 & 0.65 & 0.25 & 0.70 & 0.75 & 0.35 \\
\textbf{IPBC (ours)} & \textbf{0.80} & \textbf{0.85} & \textbf{0.50} & \textbf{0.75} & \textbf{0.80} & \textbf{0.45} & \textbf{0.88} & \textbf{0.92} & \textbf{0.60} \\
\bottomrule
\end{tabular}
\end{table*}

\begin{figure}[ht]
\centering
% Placeholders for comparative projection images
\includegraphics[width=0.95\linewidth]{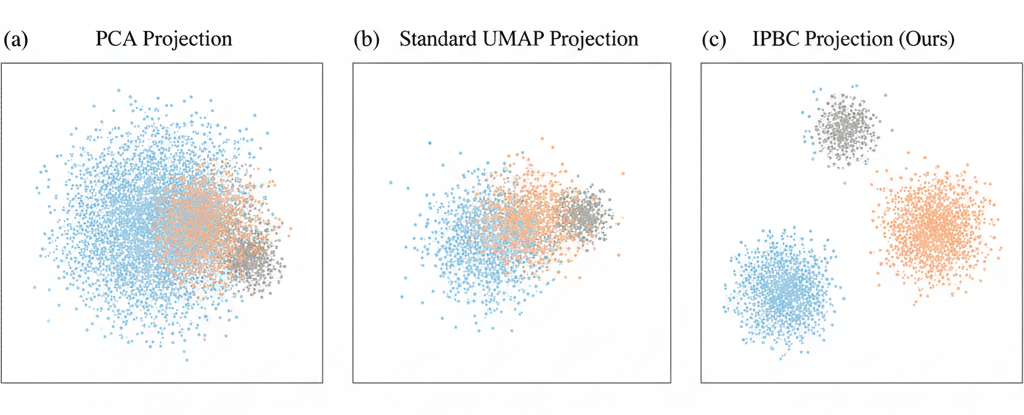}
\caption{Visual comparison of projections on MNIST. (a) PCA (poor separation). (b) Standard UMAP (good, but digits 4 and 9 are mixed). (c) Our IPBC result after 3 feedback iterations (digits 4 and 9 are now clearly separated).}
\label{fig:mnist}
\end{figure}

A qualitative illustration is shown in Figure \ref{fig:mnist}.  The PCA projection (a) shows poor separation: digit clusters overlap widely.  The standard UMAP projection (b) does much better, but some classes remain entangled (for instance, 4’s and 9’s lie in one blob).  In our IPBC projection (c), after three simulated feedback iterations the user had provided cannot-link constraints separating 4’s from 9’s. The updated projection now shows digits 4 and 9 in distinct regions.  This refined view enables clean clustering by eye. (Colors in the figure use true labels for clarity.) 

\subsection{Case Study: User Interaction Walkthrough}
We now describe a sample IPBC session on MNIST.  (1) The user loads the data and sees the initial UMAP scatterplot (without digit labels). (2) They notice that most digits form tight clusters, but the classes 4 and 9 appear to overlap. (3) Using the lasso tool, the user selects several 4’s and 9’s and marks them as a cannot-link. (4) IPBC immediately re-optimizes the projection; within a few seconds the scatterplot refreshes and 4’s and 9’s separate into different clusters. (5) The user inspects other clusters (e.g.\ 1’s, 7’s) and, if desired, adds more constraints. (6) Finally, the user clicks “Cluster,” and DBSCAN is run on the optimized layout, producing final cluster labels. In practice, each update takes only a few seconds, making the interaction smooth. This walkthrough illustrates how minimal feedback rapidly corrects a problematic cluster, turning a manual check into an efficient, guided process.

\section{Discussion} \label{sec:discussion}
Our interactive framework demonstrates the value of human--algorithm collaboration in clustering. Rather than passively running an algorithm once, IPBC treats clustering as a dialogue. The user steers the process towards the “right” solution by providing domain knowledge where the algorithm errs. This synergy leads to clusters that not only score better on quantitative metrics (Table \ref{tab:results}), but are also aligned with human understanding. In other words, IPBC makes the clustering pipeline more \emph{trustworthy}. 

\textbf{Interactive vs Static.} Conventional DR+clustering is a one-shot process: the user applies an algorithm and hopes the result is sensible. IPBC shifts this paradigm to an iterative exploration. This reflects the insight from interactive ML literature \cite{amershi2014power} that users can quickly identify and fix errors with a few adjustments. By continuously updating the projection, IPBC implements a “what-if” cycle: the user asks “what if I force these points apart?” and immediately sees the effect. This transforms clustering from a black-box output into a visual conversation.

\textbf{Domain Knowledge Integration.} IPBC provides a concrete mechanism to inject expert knowledge into unsupervised analysis. The augmented loss $L_{\rm total}$  acts as a mathematical interface: must-link constraints mathematically \emph{pull} points together, cannot-links \emph{push} points apart. This is analogous to supervised metric learning, but we apply it directly to shape the embedding. In effect, the user’s brush strokes become regularizers in the objective function. This contrasts with post-hoc adjustment methods (like flipping labels after clustering); here the feedback influences the model’s latent representation directly.

\textbf{Transparency and Trust.} Because the user sees the embedding evolve in real time and receives explicit cluster explanations, IPBC avoids being a pure black box. The iterative updates provide transparency: users can see how each constraint changes the layout. Additionally, presenting cluster characteristics in terms of original features (via decision trees) demystifies the results. We envision that this openness will increase user trust in the results, especially in sensitive domains (as urged by recent XAI research \cite{vu2022survey,hwang2022xclusters}). The user always retains control and understanding of why clusters look as they do.

\textbf{Limitations and Future Work.} The main challenge is scalability.  Re-optimizing a UMAP embedding can become slow for very large datasets (e.g.\ $N>100{,}000$).  In our prototype (thousands of points), updates took only a few seconds, but larger data will require faster implementations (e.g.\ GPU-based optimizers or incremental algorithms).  Another issue is the sensitivity of the feedback terms: choosing $\lambda_{\rm ML}, \lambda_{\rm CL}, m$ can affect convergence.  We set these empirically; future work could adapt them automatically or learn from the user’s consistency.  Finally, we assumed a simulated user; a user study will be needed to evaluate IPBC’s usability.  Extensions could include supporting other forms of feedback (e.g.\ cluster labels on a few points) or enabling 3D interactive projections.

\section{Conclusion} \label{sec:conclusion}
We have introduced IPBC, a novel interactive framework for high-dimensional clustering. By treating dimensionality reduction as a search for the “optimal viewing angle” and embedding human feedback into the projection optimization, our system enables users to iteratively refine cluster structure. The result is a win-win: clusters that score higher on standard metrics, and clusters that make sense to human experts. In summary, IPBC turns static clustering into a collaborative discovery process, bridging the gap between algorithmic analysis and human intuition.

\bibliographystyle{IEEEtran}
\bibliography{references}

\end{document}